\documentclass[10pt,twocolumn,letterpaper]{article}

\usepackage{cvpr}
\usepackage{times}
\usepackage{epsfig}
\usepackage{graphicx}
\usepackage{amsmath}
\usepackage{amssymb}

\usepackage{multirow}
\usepackage{graphicx}

\usepackage[breaklinks=true,bookmarks=false]{hyperref}

\cvprfinalcopy 

\def\cvprPaperID{3138} 

\ifcvprfinal\fi
\begin{document}

\title{Where Does It Exist: Spatio-Temporal Video Grounding for Multi-Form Sentences}

\author{Zhu Zhang$^1$, Zhou Zhao$^{1\thanks{Zhou Zhao is the corresponding author.}}$, Yang Zhao$^1$, Qi Wang$^2$, Huasheng Liu$^2$, Lianli Gao$^3$ \\
$^1$College of Computer Science, Zhejiang University\\
$^2$Alibaba Group, 
$^3$University of Electronic Science and Technology of China\\
{\tt\small \{zhangzhu, zhaozhou, zhaoyang\}@zju.edu.cn, \{wq140362, fangkong.lhs\}@alibaba-inc.com}
}

\maketitle

\begin{abstract}
In this paper, we consider a novel task, Spatio-Temporal Video Grounding for Multi-Form Sentences (STVG). Given an untrimmed video and a declarative/interrogative sentence depicting an object, STVG aims to localize the spatio-temporal tube of the queried object. STVG has two challenging settings: (1) We need to localize spatio-temporal object tubes from untrimmed videos, where the object may only exist in a very small segment of the video; (2) We deal with multi-form sentences, including the declarative sentences with explicit objects and interrogative sentences with unknown objects. Existing methods cannot tackle the STVG task due to the ineffective tube pre-generation and the lack of object relationship modeling. Thus, we then propose a novel Spatio-Temporal Graph Reasoning Network (STGRN) for this task. First, we build a spatio-temporal region graph to capture the region relationships with temporal object dynamics, which involves the implicit and explicit spatial subgraphs in each frame and the temporal dynamic subgraph across frames. We then incorporate textual clues into the graph and develop the multi-step cross-modal graph reasoning. Next, we introduce a spatio-temporal localizer with a dynamic selection method to directly retrieve the spatio-temporal tubes without tube pre-generation. Moreover, we contribute a large-scale video grounding dataset VidSTG based on video relation dataset VidOR. The extensive experiments demonstrate the effectiveness of our method. The VidSTG dataset is available at https://github.com/Guaranteer/VidSTG-Dataset.
\end{abstract}

\section{Introduction}
Grounding natural language in visual contents is a fundamental and vital task in the visual-language understanding field.
Visual grounding aims to localize the object described by the given referring expression in an image, which has attracted much attention and made great progress~\cite{hu2016natural,mao2016generation,deng2018visual,yu2018mattnet,yang2019cross}. Recently, researchers begin to explore video grounding, including temporal grounding and spatio-temporal grounding. 
Temporal sentence grounding~\cite{gao2017tall,hendricks2017localizing,zhang2019cross,xu2019multilevel} determines the temporal boundaries of events corresponding to the given sentence, but does not localize the spatio-temporal tube (i.e., a sequence of bounding boxes) of the described object. Further, spatio-temporal grounding is to retrieve the object tubes according to textual descriptions, but existing strategies~\cite{zhou2018weakly,balajee2018object,yamaguchi2017spatio,chen2019weakly} can only be applied to restricted scenarios, e.g. grounding in a frame of the video~\cite{zhou2018weakly,balajee2018object} or grounding in trimmed videos~\cite{yamaguchi2017spatio,chen2019weakly}.
Moreover, due to the lack of bounding box annotations, researchers~\cite{zhou2018weakly,chen2019weakly} can only adopt a weakly-supervised setting, leading to suboptimal performance.

\begin{figure}[t]
\centering
\includegraphics[width=0.47\textwidth]{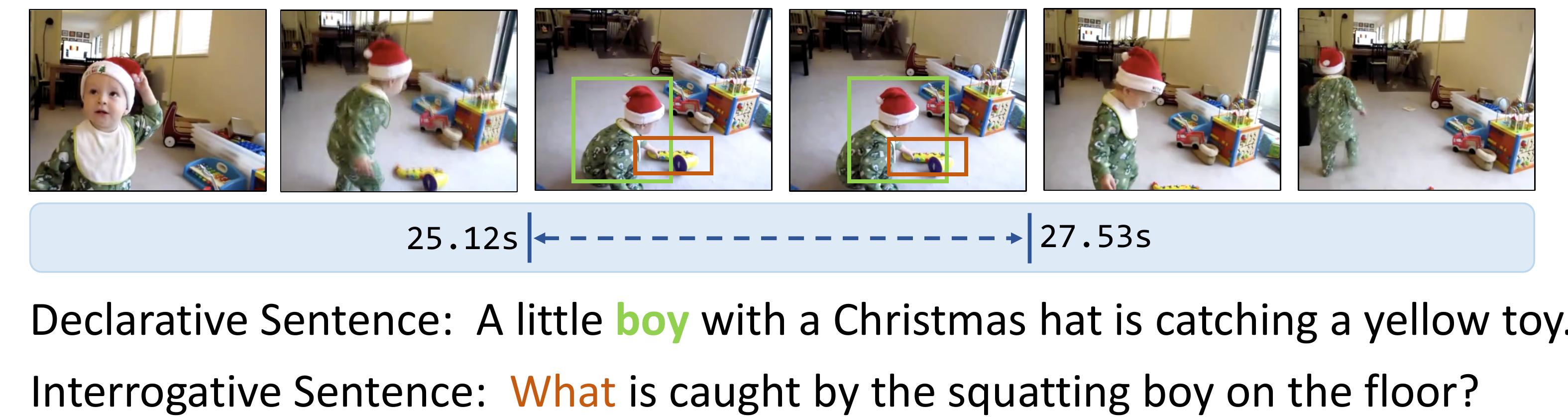}
\caption{An example of STVG for multi-form sentences.}\label{fig:exp}
\end{figure}

To break through above restrictions, we propose a novel task, Spatio-Temporal Video Grounding for Multi-Form Sentences (STVG). Concretely, as illustrated in Figure~\ref{fig:exp}, given an untrimmed video and a declarative/interrogative sentence depicting an object, STVG aims to localize the spatio-temporal tube of the queried object.

Compared with previous video grounding~\cite{zhou2018weakly,balajee2018object,yamaguchi2017spatio,chen2019weakly}, STVG has two novel and challenging points. First, we localize spatio-temporal object tubes from untrimmed videos. The objects may exist in a very small segment of the video and be hard to distinguish. And the sentences may only describe a short-term state of the queried object, e.g. the action "catching a toy" of the boy in Figure~\ref{fig:exp}. So it is crucial to determine the temporal boundaries of object tubes by sufficient cross-modal understanding.
Secondly, STVG deals with multi-form sentences, that it, not only grounds the conventional declarative sentences with explicit objects, but also localizes the interrogative sentences with unknown objects, for example, the sentence "What is caught by the squatting boy on the floor?" in Figure~\ref{fig:exp}.
Due to the lack of the explicit characteristics of objects (e.g. the class "toy" and visual appearance "yellow"), grounding for interrogative sentences can only depend on relationships between the unknown object and other objects (e.g. the action relation "caught by the squatting boy" and spatial relation "on the floor").
Thus, sufficient relationship construction and cross-modal relation reasoning are crucial for the STVG task.

Existing video grounding methods~\cite{yamaguchi2017spatio,chen2019weakly} often extract a set of spatio-temporal tubes from trimmed videos and then identify the target tube that matches the sentence. 
However, this framework may be unsuitable for STVG. On the one hand, the performance of this framework is heavily dependent on the quality of tube candidates. But it is difficult to pre-generate high-quality tubes without textual clues, since the sentences may describe a short-term state of objects in a very small segment, but the existing tube pre-generation framework~\cite{yamaguchi2017spatio,chen2019weakly} can only produce the complete object tubes from trimmed videos. On the other hand, these methods only consider single tube modeling and ignore the relationships between objects. However, object relations are vital clues for the STVG task, especially for interrogative sentences that may only offer the interactions of the unknown objects with other objects. Thus, these approaches cannot deal with the complicated grounding of STVG.

To tackle above problems, we propose a novel Spatio-Temporal Graph Reasoning Network (STGRN) to capture region relationships with temporal object dynamics and directly localize the spatio-temporal tubes without tube pre-generation. Specifically, we first parse the video into a spatio-temporal region graph. Existing visual graph modeling approaches~\cite{yao2018exploring,li2019relation} often build the spatial graph in an image, which cannot utilize the temporal dynamics information in videos to distinguish the subtle differences of object actions, e.g. distinguish "open the door" and "close the door". Different from them, our spatio-temporal region graph not only involves the implicit and explicit spatial subgraphs in each frame, but also includes a temporal dynamic subgraph across frames. The spatial subgraphs can capture the region-level relationships by implicit or explicit semantic interactions, and the temporal subgraph can incorporate the dynamics and transformation of objects across frames to further improve the relation understanding. Next, we fuse the textual clues into this spatio-temporal graph as the guidance, and develop the multi-step cross-modal graph reasoning. The multi-step process can support the multi-order relation modeling like "a man hug a baby wearing a red hat". After it, we introduce a spatio-temporal localizer to directly retrieve the spatio-temporal object tubes from the region level. Concretely, we first employ a temporal localizer to determine the temporal boundaries of the tube, and then apply a spatial localizer with a dynamic selection method to ground the object in each frame and generate a smooth tube.

To facilitate this STVG task, we contribute a large-scale video grounding dataset VidSTG by adding the multi-form sentence annotations into video relation dataset VidOR. 

Our main contributions can be summarized as follows:
\begin{itemize}
\item We propose a novel task STVG to explore the spatio-temporal video grounding for multi-form sentences.
\item We develop a novel STGRN to tackle this STVG task, which builds a spatio-temporal graph to capture the region relationships with temporal object dynamics, and employs a spatio-temporal localizer to directly retrieve the spatio-temporal tubes without tube pre-generation.
\item We contribute a large-scale video grounding dataset VidSTG as the benchmark of the STVG task.
\item The extensive experiments show the effectiveness of our proposed STGRN method.
\end{itemize}

\begin{figure*}[t]
\centering
\includegraphics[width=0.98\textwidth]{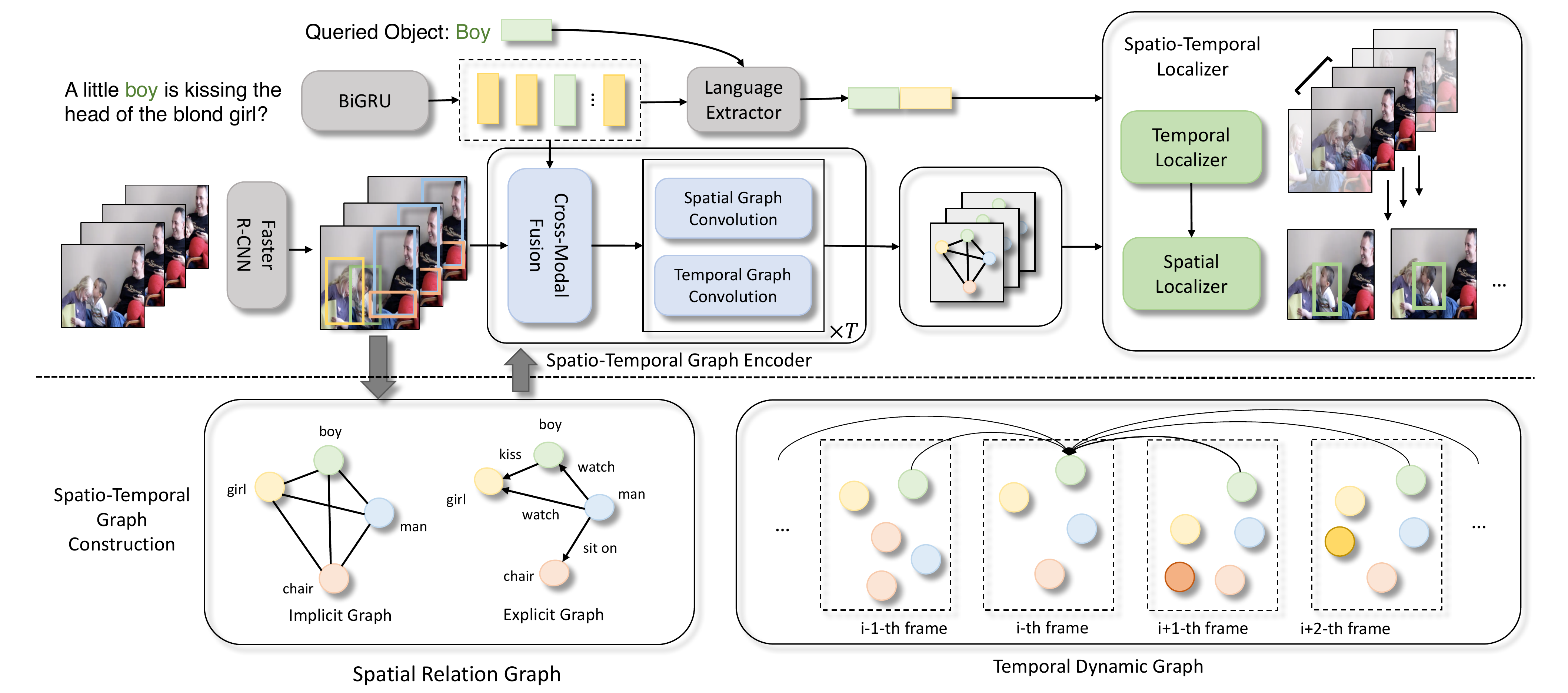}
\caption{The Overall Architecture of the Spatio-Temporal Graph Reasoning Network (STGRN). We first extract the region features and learn the query representation. Next, we apply a spatio-temporal graph encoder to develop the multi-step cross-modal graph reasoning. After it, a spatio-temporal localizer with $T$ spatio-temporal convolution layers directly retrieves the tubes from the region level. }\label{fig:framework}
\end{figure*}

\section{Related Work}

\subsection{Temporal Localization via Natural Language}
Temporal natural language localization is to detect the video clip depicting the given sentence.
Early approaches~\cite{hendricks2017localizing,gao2017tall,hendricks2018localizing,liu2018attentive,liu2018cross} are mainly based on the sliding window framework, which first samples abundant candidate clips and then ranks them. Recently, people begin to develop holistic and cross-modal methods~\cite{zhang2019man,chen2018temprally,chen2019localizing,wang2019language,zhang2019cross,xu2019multilevel} to solve this problem.
Chen, Zhang and Xu et al.~\cite{chen2018temprally,chen2019localizing,zhang2019cross,lin2020moment,xu2019multilevel} build frame-by-word interactions from visual and textual contents to aggregate the matching clues.
Zhang et al.~\cite{zhang2019man} deal with the structural and semantic misalignment challenges by an explicitly structured graph. 
Wang et al.~\cite{wang2019language} propose a reinforcement learning framework to adaptively observe frame sequences and associate video contents with sentences.
Further, Mithun and Lin et al.~\cite{mithun2019weakly,lin2019weakly} devise the weakly-supervised temporal localization methods requiring only coarse video-level annotations for training.
And besides the natural language query, Zhang et al.~\cite{zhang2019localizing} attempt to localize the unseen temporal clip according to an image query.
Although these methods have achieved promising performance, they still remain in the temporal grounding. We further explore spatio-temporal grounding in this paper.

\subsection{Object Grounding in Images/Videos}
Visual grounding~\cite{hu2016natural,mao2016generation,xiao2017weakly,yu2016modeling,nagaraja2016modeling,yu2017joint,deng2018visual,zhuang2018parallel,hu2017modeling,yu2018mattnet,zhang2018grounding,yang2019cross,yang2019dynamic} aims to localize the visual object described by the given referring expression. Early methods~\cite{hu2016natural,mao2016generation,xiao2017weakly,yu2016modeling,nagaraja2016modeling,yu2017joint} often extract object features by CNN, model the language expression through RNN and learn the object-language matching. Some recent approaches~\cite{yu2018mattnet,hu2017modeling} decompose the expression into multiple components and calculate matching scores of each module. And Deng and Zhuang et al.~\cite{deng2018visual,zhuang2018parallel} apply the co-attention mechanism to build cross-modal interactions. Further, Yang et al.~\cite{yang2019dynamic,yang2019cross} explore the relationships between objects to improve the accuracy.
As for video grounding, existing works~\cite{zhou2018weakly,balajee2018object,yamaguchi2017spatio,chen2019weakly,yamaguchi2017spatio,chen2019weakly} can only be applied to restricted scenarios.
Zhou and Balajee et al.~\cite{zhou2018weakly,balajee2018object} only ground natural language in a frame of the video.
With a sequence of transcriptions and temporal alignment video clips, Huang and Shi et al.~\cite{Huang_2018_CVPR,shi2019not} ground nouns or pronouns in specific frames by the weakly-supervised MIL methods.
And Chen and Yamaguchi et al.~\cite{yamaguchi2017spatio,chen2019weakly} localize spatio-temporal object tubes from trimmed videos, but cannot directly deal with raw video streams. 
In this paper, we propose a novel STVG task to further explore the spatio-temporal video grounding for multi-form sentences.

\section{The Proposed Method}
Given a video ${v} \in V$ and a sentence ${s} \in S$ depicting an object, STVG is to retrieve its spatio-temporal tube. Figure~\ref{fig:framework} illustrates the overall framework.

\subsection{Video and Text Encoder}
We first apply a pre-trained Faster R-CNN~\cite{ren2015faster} to extract a set of regions for each frame, where a video contains $N$ frames and the $t$-th frame corresponds to $K$ regions denoted by $\{{r}^{t}_i\}_{i=1}^K$. A region ${r}^{t}_i$ is associated with its visual feature ${\bf r}^{t}_i \in  \mathbb{R}^{d_r}$ and bounding box vector ${\bf b}^{t}_i = [x^t_i, y^t_i, w^t_i, h^t_i]$, where $(x^t_i, y^t_i)$ are the normalized coordinates of the center point and $(w^t_i, h^t_i)$ are the normalized width and height. Besides, we obtain the frame features $\{{\bf f}^t\}_{t=1}^N$, i.e., the visual feature of the entire frame from Faster R-CNN.

For the sentence $s$ with $L$ words, we first input the word embeddings into a bi-directional GRU~\cite{chung2014empirical} to learn the word semantic features $\{{\bf s}_i\}_{i=1}^L \in \mathbb{R}^{d_s}$, where ${\bf s}_i$ is the concatenation of the forward and backward hidden states of step $i$. As the STVG grounds objects described in the declarative or interrogative sentences, we need to extract the query representation from language context. First, we select the entity feature ${\bf s}^e$ from $ \{{\bf s}_i\}_{i=1}^L$, which represents the queried object, for example, ${\bf s}_3$ for "boy" in Figure~\ref{fig:framework}. Note that for interrogative sentences, the feature of "who" or "what" is chosen as the entity feature. Next, an attention method aggregates the textual clues from language context by
\begin{eqnarray}
\begin{aligned}
&\gamma_i = {({\bf W}_1^s{\bf s}^{e})}^{\top}\cdot ({\bf W}_2^s{\bf s}_{i}), \ {\widetilde \gamma_i} = \frac{{\rm exp}(\gamma_i)}{\sum_{i=1}^{L} {\rm exp}(\gamma_i)},  \\
&{\bf s}^a = \sum_{i=1}^{L} {\widetilde \gamma_i} {\bf s}_{i}, \ {\bf s}^q = [{\bf s}^e; {\bf s}^a],
\end{aligned}
\end{eqnarray}
where ${\bf W}_1^s$ and ${\bf W}_2^s$ are projection matrices, ${\bf s}^a$ is the entity-aware feature and ${\bf s}_q$ is the query representation.

\subsection{Spatio-Temporal Graph Encoder}
Our STVG task requires to capture the object relationships and develop the cross-modal understanding, especially for interrogative sentences that may only offer the interaction information of the unknown object with other objects. Thus, we build a spatio-temporal region encoder with $T$ spatio-temporal convolution layers to capture the region relationships with temporal object dynamics and support the multi-step cross-modal graph reasoning. 

\subsubsection{Graph Construction}
We first parse the video into a spatio-temporal region graph, which involves the implicit spatial subgraph ${\mathcal G}_{imp} = ({\mathcal V},{\mathcal E}_{imp})$ , explicit spatial subgraph ${\mathcal G}_{exp} = ({\mathcal V},{\mathcal E}_{exp})$ in each frame and temporal dynamic subgraph ${\mathcal G}_{tem} = ({\mathcal V},{\mathcal E}_{tem})$  across frames. 
Three subgraphs all treat regions as their vertexes ${\mathcal V}$ but have different edges. Note that we add the self-loop of each vertex in each subgraph.

\textbf{Implicit Spatial Graph.} We regard the fully-connected region graph in each frame as the implicit spatial graph ${\mathcal G}_{imp}$, where ${\mathcal E}_{imp}$ contains $K \times K$ undirected and unlabeled edges in each frame (including self-loops). 

\textbf{Explicit Spatial Graph.} We extract the region triplet $\langle r_i^t, p^t_{ij}, r_j^t \rangle$ to construct the explicit spatial graph, where $r_i^t$ and $r_j^t$ are the i-th and j-th regions in frame t, and $p^t_{ij}$ is the relation predicate between them. Each triplet can be regarded as an edge from $i$ to $j$.
Thus, explicit graph construction can be formulated as a relation classification task~\cite{yao2018exploring,zellers2018neural}.
Concretely, given the feature $[{\bf r}_i^t ; {\bf b}_i^t]$ of region $i$, feature $[{\bf r}_j^t ; {\bf b}_j^t]$ of region $j$ and the united feature $[{\bf r}_{ij}^t ; {\bf b}_{ij}^t]$ of the union bounding box of $i$ and $j$ (also extracted by Faster R-CNN), we first transform three features via different linear leyers and then concatenate them into a classification layer to predict the relations. Similer to existing works~\cite{yao2018exploring,li2019relation}, we train such a classifier on the Visual Genome dataset~\cite{krishna2017visual}, where we select top-50 frequent predicates in its training data and add an extra $no\_relation$ class for non-existent edges.
We then predict the relationships between $r_i^t$ and $r_j^t$. Eventually, the edges ${\mathcal E}_{exp}$ have 3 directions (including $i$-to-$j$, $j$-to-$i$ and $i$-to-$i$ of self-loops) and 51 types of labels (top-50 classes plus the self-loop).

\textbf{Temporal Dynamic Graph.} 
While the spatial graphs model region-region interactions, our temporal graph is to capture the dynamics and transformation of objects across frames. So we expect to connect the regions containing the same object in different frames and then learn more expressive and discriminative object features. For the frame $t$, we connect its regions with adjacent $2M$ frames ($M$ for forward frames and $M$ for backward). Too distant frames cannot provide the real-time dynamics. Concretely, we first define the linking score $s(r_i^t, r_j^{k})$ between $r_i^t$ from frame $t$ and $r_j^{k}$ from frame $k$ by
\begin{eqnarray}
& s(r_i^t, r_j^{k}) = {\rm cos}({\bf r}_i^t, {\bf r}_j^{k}) + \frac{\epsilon}{|k-t|} \cdot {\rm IoU}(r_i^t, r_j^{k}),
\end{eqnarray}
where ${\rm cos}(\cdot)$ is the cosine similarity of two features, ${\rm IoU}(\cdot)$ is the intersection-over-union of two regions, and $\epsilon$ is the balanced scalar. Here we simultaneously consider the appearance similarity and spatial overlap ratio of two regions. And the temporal distance $|k-t|$ of two frames is used to limit the ${\rm IoU}$ score, that is, for distant frames, the linking score is mainly determined by feature similarity rather than the spatial overlap. Next, for the region ${r}_i^t$, we select the region $r_j^{k}$ with the maximal linking score from frame $k$ to build an edge, and get $2M+1$ edges for each region (including the self-loop). The unlabeled temporal edges ${\mathcal E}_{tem}$ have 3 directions: forward, backward and self-loop.

\subsubsection{Multi-Step Cross-Modal Graph Reasoning}
After graph construction, we incorporate the textual clues into this graph and develop the multi-step cross-modal graph reasoning by $T$ spatio-temporal convolution layers.

\textbf{Cross-Modal Fusion.} To capture the relationships associated with the sentence, we first use a cross-modal fusion that dynamically injects textual evidences into the spatio-temporal graph. Concretely, we first 
utilize an attention mechanism to aggregate the words features for each region. For a region $r_i^t$, we calculate the attention weights over word features $\{{\bf s}_i\}_{i=1}^L$, denoted by
\begin{eqnarray}
\begin{aligned}
& \alpha^t_{ij} =  {\bf w}^{\top}_m{\rm tanh}({\bf W}^{m}_{1} {\bf r}_{i}^{t}+ {\bf W}^{m}_{2}{\bf s}_{j}+{\bf b}^{m}), \\
& {\widetilde \alpha^t_{ij}} = \frac{{\rm exp}(\alpha^t_{ij})}{\sum_{j=1}^{L} {\rm exp}(\alpha^t_{ij})}, \ {\bf q}^{t}_{i} = \sum_{j=1}^{L} {\widetilde \alpha^t_{ij}} {\bf s}_{j},
\end{aligned} 
\end{eqnarray}
where $ {\bf W}_{1}^{m}$, $ {\bf W}_{2}^{m}$ are projection matrices, ${\bf b}^{m}$ is the bias and ${\bf w}^{\top}_m$ is the row vector. And ${\bf q}^{t}_{i}$ is the region-aware textual feature for each region $i$ in frame t.

Next, we build the textual gate that takes language information as the guidance to weaken the text-irrelevant regions, given by
\begin{eqnarray}
&{\bf g}_i^t = \sigma({\bf W}^g{\bf q}^t_i + {\bf b}^g), \ {\bf v}_i^t = [{\bf r}_i^t \odot {\bf g}_i^t ; {\bf q}^t_i],
\end{eqnarray}
where $\sigma$ is the sigmoid function, $\odot$ is the element-wise multiplication, ${\bf g}_i^t \in \mathbb{R}^{d_r}$ means the textual gate for region $r_i^t$. And we then concatenate the filtered region feature and textual feature to obtain the cross-modal region features $\{\{{\bf v}_i^t\}_{i=1}^K\}_{t=1}^N$. Next, we develop $T$ spatio-temporal convolution layers for the multi-step graph reasoning.

\textbf{Spatial Graph Convolution.} In each layer, we first develop the spatial graph convolution to capture visual relationships among regions in each frame. Concretely, with cross-modal region features $\{\{{\bf v}_i^t\}_{i=1}^K\}_{t=1}^N$,  we first adopt the implicit graph convolution on ${\mathcal G}_{imp}$ that is undirected and unlabeled, given by
\begin{eqnarray}
\begin{aligned}
& {\bf \overline v}_i^t = \sum \limits_{j \in {\mathcal N}_i(r^t_i)} \alpha_{ij}^{imp} \cdot ({\bf W}^{imp} {\bf  v}_j^t ), \\
& \alpha_{ij}^{imp} = \frac{{\rm exp}({({\bf U}^{imp}[{\bf  v}_{i}^t;{\bf b}_{i}^t ])}^{\top}\cdot ({\bf U}^{imp}[{\bf  v}_{j}^t;{\bf b}_{j}^t ]))}{\sum \limits_{j \in {\mathcal N}_i(r^t_i)} {\rm exp}({({\bf U}^{imp}[{\bf  v}_{i}^t;{\bf b}_{i}^t ])}^{\top}\cdot ({\bf U}^{imp}[{\bf  v}_{j}^t;{\bf b}_{j}^t ]))},
\end{aligned}
\end{eqnarray}
where ${\mathcal N}_i(r^t_i)$ are the regions connected with $r_i^t$ in ${\mathcal G}_{imp}$. The implicit graph convolution can be regarded as a variant of self-attention and we compute the coefficient $\alpha_{ij}^{imp}$ by combining the visual features and region locations. 

Simultaneously, we develop the explicit graph convolution.  Different from the original undirected GCN~\cite{kipf2016semi,velivckovic2017graph}, we consider the direction and label information of edges on the directed and labeled ${\mathcal G}_{exp}$, given by 
\begin{eqnarray}
& {\bf \hat v}_i^t = \sum \limits_{j \in {\mathcal N}_e(r^t_i)} {\alpha}^{exp}_{lab(i,j)} \cdot ({\bf W}^{exp}_{dir(i,j)} {\bf  v}_j^t + {\bf b}^{exp}_{lab(i,j)}),
\end{eqnarray}
where ${\bf W}^{exp}_{(\cdot)}$ are optional matrices by the direction $dir(i,j)$ of edge $(i,j)$,  ${\bf b}^{exp}_{(\cdot)}$ are optional bias by the label of edge $(i,j)$. Here, the edge has three directions ($i$-to-$j$, $j$-to-$i$, $i$-to-$i$) and 51 types. ${\mathcal N}_e(r^t_i)$ are the regions connected with $r_i^t$. Moreover, the relation coefficient ${\alpha}^{exp}_{(\cdot)}$ can also be chosen by the label of edge $(i,j)$. Different sentences describe different relations and their grounding is heavily dependent on the specific relation understanding. Thus, the coefficient of explicit edges can be decided by query presentation ${\bf s}^q$, given by
\begin{eqnarray}
& {\alpha}^{exp} = {\rm Softmax}({\bf W}^r{\bf s}^q+{\bf b}^r),
\end{eqnarray}
where ${\alpha}^{exp} \in \mathbb{R}^{51}$ corresponds to the coefficients of 51 types of relationships.

\textbf{Temporal Graph Convolution.} We next develop the temporal graph convolution on the directed and unlabeled graph ${\mathcal G}_{tem}$ to capture the dynamics and transformation of objects across frames. We consider the forward, backward and self-loop edges for each region ${r}_i^t$ with the cross-modal feature ${\bf v}_i^t$, denoted by
\begin{eqnarray}
\begin{aligned}
& {\bf \widetilde v}_i^t = \sum \limits_{j \in {\mathcal N}_t(r^t_i)} \alpha_{ij}^{tem} \cdot {\bf W}^{tem}_{dir(i,j)} {\bf v}_j,\\
& \alpha_{ij}^{tem} = \frac{{\rm exp}({({\bf U}^{tem}{\bf v}_{i}^t)}^{\top}\cdot ({\bf V}^{tem}_{dir(i,j)}{\bf v}_{j}))}{\sum_{j \in {\mathcal N}_t(r^t_i)} {\rm exp}({({\bf U}^{tem}{\bf v}_{i}^t)}^{\top}\cdot ({\bf V}^{tem}_{dir(i,j)}{\bf v}_{j}))},
\end{aligned}
\end{eqnarray}
where ${\bf W}^{tem}_{(\cdot)}$ and ${\bf V}^{tem}_{(\cdot)}$ are matrices and ${dir(i,j)}$ indicates the direction of edge $(i,j)$ to select the corresponding projection matrix, where the temporal edges have three directions. And $\alpha_{ij}^{tem}$ is the semantic coefficient for each neighborhood region.

Next, we combine the outputs of spatial and temporal graph convolutions and obtain the result ${\bf v}^{t(1)}_i$ of the first spatio-temporal convolution layer by 
\begin{eqnarray}
& {\bf v}^{t(1)}_i = {\rm ReLU}({\bf \overline v}_i^t + {\bf \hat v}_i^t + {\bf \widetilde v}_i^t  + {\bf v}_i^t ).
\end{eqnarray}
In order to support multi-order relation modeling, we perform the multi-step encoding by the spatio-temporal graph encoder with $T$ spatio-temporal convolution layers and learn final relation-aware region features $\{\{{\bf m}_i^t\}_{i=1}^K\}_{t=1}^N$ .

\subsection{Spatio-Temporal Localizer}
In this section, we devise a spatio-temporal localizer to determine the temporal tube boundaries and spatio-temporal tubes of objects from the region level.

\textbf{Temporal Localizer.} We first introduce the temporal localizer, which estimates a set of candidate clips and adjust their boundaries to obtain the temporal grounding~\cite{zhang2019cross}. Specifically, we first aggregate the relation-aware region graph into the frame level by an attention mechanism. With the query representation ${\bf s}^q$, the region features of each frame are attended  by 
\begin{eqnarray}
\begin{aligned}
& \beta^t_{i} = {\bf w}^{\top}_f{\rm tanh}({\bf W}^{f}_{1} {\bf m}_{i}^{t}+ {\bf W}^{f}_{2}{\bf s}_{q}+{\bf b}^{f}), \\
&{\widetilde \beta^t_{i}} = \frac{{\rm exp}(\beta^t_{i} )}{\sum_{i=1}^{K} {\rm exp}(\beta^t_{i} )}, \ {\bf m}^{t} = \sum_{i=1}^{K}{\widetilde \beta^t_{i}} {\bf m}_{i}^t,
\end{aligned}
\end{eqnarray}
where ${\bf m}^t$ represents the relation-aware feature of frame $t$. We then concatenate these features with their corresponding global frame features $\{{\bf f}^t\}_{t=1}^N$, and apply another BiGRU to learn final frame features $\{{\bf h}^t\}_{t=1}^N$. Next, we define multi-scale candidate clips at each time step $t$ as  $B^t =\{ ({ s}_{i}^t, { e}_{i}^t) \}_{i=1}^P$, where $({ s}_{i}^t, {e}_{i}^t)  = (t - w_i/2, t + w_i/2)$ are the start and end boundaries of the $i$-th clips, $w_i$ is the temporal length of $i$-th clip and $P$ is the clip number. After it, we estimate all candidate clips by a linear layer with the sigmoid nonlinearity and simultaneously produce the offsets of their boundaries, given by
\begin{eqnarray}
& {\bf C}^{t} = \sigma({\bf W}^c [{\bf h}^t;{\bf s}^q] + {\bf b}^c), \  {\bf \delta}^{t} = {\bf W}^o [{\bf h}^t ;{\bf s}^q]+ {\bf b}^o,
\end{eqnarray}
where ${\bf C}^{t} \in \mathbb{R}^{P}$ corresponds to confidence scores of $P$ candidates at step $t$ and ${\delta}^{t} \in \mathbb{R}^{2P}$ are the offsets of $P$ clips.

The temporal localizer has two losses: the alignment loss for the clip selection and a regression loss for boundary adjustments. Concretely, for alignment loss, we first compute the temporal IoU score ${\hat C}_{i}^t$ of each clip with the ground truth. 
And the alignment loss is denoted by 
\begin{equation}
{\mathcal L}_{align}  = -\frac{1}{NP}\sum_{t=1}^{N}\sum_{i=1}^{P}(1 - {\hat C}_{i}^t) \cdot {\rm log}(1 - { C}_{i}^t) + {\hat C}_{i}^t \cdot {\rm log}({C}_{i}^t),
\end{equation}
where we use the temporal IoU score ${\hat C}_{i}^t$ rather than 0/1 score to further distinguish high-score clips. Next, we fine-tune the boundaries of the best clip with highest ${C}_{i}^t$, which has the boundaries $({s}, {e})$ and offsets $({\delta}_s, {\delta}_e)$. We first compute the offsets of this clip from ground truth boundaries $({\hat s}, {\hat e})$ by ${\hat \delta}_s = s - {\hat s}$ and ${\hat \delta}_e = e - {\hat e}$ and define the regression loss by 
\begin{eqnarray}
& {\mathcal L}_{reg} = {\rm R}({\delta}_s - {\hat \delta}_s) + {\rm R}({\delta}_e - {\hat \delta}_e),
\end{eqnarray}
where ${\rm R}$ represents the smooth L1 function.

\textbf{Spatial Localizer.} With the temporal grounding, we next localize the target regions in each frame. For the $t$-th frame with region features $\{{\bf m}_i^t\}_{i=1}^K$, we directly estimate the matching scores of each region by integrating the query representation ${\bf s}^q$ and final frame feature ${\bf h}^t$, denoted by
\begin{eqnarray}
& {S}^{t}_i = \sigma({\bf W}^c [{\bf m}^t_i;{\bf s}^q;{\bf h}^t] + {\bf b}^c),
\end{eqnarray}
where ${S}^{t}_i$ is the matching score of region $i$ of frame $t$. Similar to temporal alignment loss, the spatial loss first compute the spatial IoU score ${\hat S}_{i}^t$ for each region with the ground truth region, where the frames outside the temporal ground truth are omitted. And the spatial loss is denoted by 
\begin{equation}
{\mathcal L}_{exp}  = -\frac{1}{K|{\mathcal S}_t|}\sum_{t \in {\mathcal S}_{t}}\sum_{i=1}^{K}(1 - { \hat S}_{i}^t) \cdot {\rm log}(1 - {S}_{i}^t) + { \hat S}_{i}^t \cdot {\rm log}({S}_{i}^t), 
\end{equation}
where ${\mathcal S}_t$ is the set of frames in the temporal ground truth.

Eventually, we devise a multi-task loss to train our proposed STGRN in an end-to-end manner, given by
\begin{eqnarray}
{\mathcal L}_{STGRN} = {\lambda}_1 {\mathcal L}_{align} +  {\lambda}_2 {\mathcal L}_{reg} + {\lambda}_3 {\mathcal L}_{exp},
\end{eqnarray}
where ${\lambda}_1$, ${\lambda}_2$ and ${\lambda}_3$ are the hyper-parameters to control the balance of three losses.

\subsection{Dynamic Selection Method}
During inference, we first retrieve the temporal boundaries $({\rm T}_s, {\rm T}_e)$ of the tube from the temporal localizer and then determine the grounded region for each frame by the spatial localizer.
A greedy method directly selects the regions with highest matching scores ${S}^{t}_i$. However, such generated tubes may not be very smooth. The bounding boxes between adjacent frames may have too large displacements. Thus, to make the trajectory smoother, we introduce a dynamic selection method. Concretely, we first define the linking score $s(r_i^t, r_j^{t+1})$ between regions of successive frames $t$ and $t+1$ by
\begin{eqnarray}
& s(r_i^t, r_j^{t+1}) = {S}^{t}_i + {S}^{t+1}_i + {\theta} \cdot {\rm IoU}(r_i^t, r_j^{t+1}),
\end{eqnarray}
where ${S}^{t}_i$ and ${S}^{t+1}_i$ are matching scores of regions $r_i^t$ and $r_i^{t+1}$, and $\theta$ is the balanced scalar which is set to 0.2. Next, we generate the final spatio-temporal tube ${\rm Y}$ with the maximal energy $E$ given by
\begin{eqnarray}
& E({\rm Y}) = \frac{1}{|{\rm T}_e - {\rm T}_s|}\sum_{t = {\rm T}_s}^{{\rm T}_e - 1}s(r_i^t, r_j^{t+1}),
\end{eqnarray}
where $({\rm T}_s, {\rm T}_e)$ are the temporal boundaries and we solve this optimization problem using a Vitervi algorithm~\cite{gkioxari2015finding}.

\begin{table}[t]
\centering
\caption{Dataset Statistics about the Number of Declarative and Interrogative Sentences.}\label{table:dataset}
\begin{tabular}{c|ccc}
\hline
     & \#Declar. Sent. & \#Inter. Sent.&All \\
\hline
Training&36,202&44,482&80,684\\
Validation&3,996&4,960&8,956\\
Testing&4,610&5,693&10,303\\
\hline
All&44,808&55,135&99,943\\
\hline
\end{tabular}
\end{table}

\begin{table*}[t]
\centering
\caption{Performance Evaluation Results on the VidSTG Dataset.}\label{table:mainexp}
\begin{tabular}{c|cccc|cccc}
\hline
\multirow{2}{*}{Method}& \multicolumn{4}{c|}{Declarative Sentence Grounding} & \multicolumn{4}{c}{Interrogative Sentence Grounding} \\
 & m\_tIoU & m\_vIoU&   vIoU@0.3 &vIoU@0.5 &m\_tIoU & m\_vIoU&   vIoU@0.3 &vIoU@0.5  \\
\hline
Random&5.18\%&0.69\%&0.04\%&0.01\%&5.35\%&0.60\%&0.02\%&0.01\%\\
\hline
GroundeR + TALL&\multirow{3}{*}{34.63\%}&9.78\%&11.04\%&4.09\%&\multirow{3}{*}{33.73\%}&9.32\%&11.39\%&3.24\% \\
STPR + TALL&&10.40\%&12.38\%&4.27\%&&9.98\%&11.74\%&4.36\% \\
WSSTG + TALL&&11.36\%&14.63\%&5.91\%&&10.65\%&13.90\%&5.32\%\\
\hline
GroundeR + L-Net&\multirow{3}{*}{40.86\%}&11.89\%&15.32\%&5.45\%&\multirow{3}{*}{39.79\%}&11.05\%&14.28\%&5.11\% \\
STPR + L-Net&&12.93\%&16.27\%&5.68\%&&11.94\%&14.73\%&5.27\%\\
WSSTG + L-Net&&14.45\%&18.00\%&7.89\%&&13.36\%&17.39\%&7.06\%\\
\hline   
STGRN (Greedy)&\multirow{2}{*}{\bf 48.47\%}&18.99\%&23.63\%&13.48\%&\multirow{2}{*}{\bf 46.98\%}&17.46\%&20.02\%&11.92\%\\ 
STGRN&&{\bf 19.75\%}&{\bf 25.77\%}&{\bf 14.60\%}&&{\bf 18.32\%}&{\bf 21.10\%}&{\bf 12.83\%}\\
\hline
\hline
GroundeR + Tem. Gt& -&28.80\%&43.20\%&22.74\%&-&26.11\%&38.37\%&18.34\%\\
STPR + Tem. Gt& -&29.72\%&44.78\%&23.83\%&-&26.97\%&39.89\%&20.07\%\\
WSSTG + Tem. Gt&-&33.32\%&50.01\%&29.98\%&-&30.05\%&44.54\%&25.76\%\\
STGRN + Tem. Gt& - &{\bf 38.04\%}&{\bf 54.47\%}&{\bf 34.80\%}&-&{\bf 35.70\%}&{\bf 47.79\%}&{\bf 31.41\%}\\
\hline
\end{tabular}
\end{table*}

\section{Dataset}
As a novel task, STVG lacks a suitable dataset as the benchmark. 
Therefore, we contribute a large-scale spatio-temporal video grounding dataset VidSTG by augmenting the sentence annotations on VidOR~\cite{shang2019annotating}. 

\subsection{Dataset Annotation}
VidOR~\cite{shang2019annotating} is the existing largest object relation dataset, containing 10,000 videos and fine-grained annotations for objects and their relations. Specifically, VidOR annotates 80 categories of objects with dense bounding boxes and annotates 50 categories of relation predicates among objects (8 spatial relations and 42 action relations). Specifically, VidOR denotes a relation as a triplet $\langle subject, predicate, object \rangle$ and each triplet is associated with the temporal boundaries and spatio-temporal tubes of $subject$ and $object$. Based on VidOR, we can select the suitable triplets, and describe the $subject$ or $object$ with multi-form sentences. 
Taking VidOR as the basic dataset has many advantages. On the one hand, we can avoid labor-intensive annotations for bounding boxes. On the other hand, the relationships in the triplets can be simply incorporated into the annotated sentences.

We first split and clean the VidOR data, and then annotate the rest video-triplet pairs with multi-form sentences. The cleaning process is introduced in the supplementary material. 
For each video-triplet pair, we choose the $subject$ or $object$ as the queried object, and then describe its appearance, relationships with other objects and visual environments. For interrogative annotations, the appearance of queried objects is ignored. We discard video-triplet pairs that are too hard to give a precise description. And a video-triplet pair may correspond to multiple sentences. 

\subsection{Dataset Statistics}
After annotation, there are 99,943 sentence descriptions about 79 types of queried objects for 44,808 video-triplet pairs, shown in Table~\ref{table:dataset}. 
The average duration of videos is 28.01s and the average temporal length of object tubes is 9.68s. The average lengths of declarative and interrogative sentences are 11.12 and 8.98, respectively. 
Further, we provide the distribution of 79 types of queried objects and some annotation examples in the supplementary material.

\section{Experiments}
\subsection{Experimental Settings}

\textbf{Implementation Details.} In STGRN, we first sample 5 frames per second and downsample the frame number of overlong videos to 200. We then pretrain the Faster R-CNN on MSCOCO~\cite{lin2014microsoft} to extract 20 region proposals for each frame (i.e. $K$ = 20). The region feature dimension $d_r$ is 1,024 and we map it 256 before graph modeling. For sentences, we use a pretrained Glove word2vec~\cite{pennington2014glove} to extract 300-d word embeddings. As for the hyper-parameters, we set $M$ to 5, $\epsilon$ to 0.8, $\theta$ to 0.2 and set $\lambda_1$,  $\lambda_2$,  $\lambda_3$ to 1.0, 0.001 and 1.0, respectively. The layer number $T$ of the spatio-temporal graph encoder is set to 2. For the temporal localizer, we set $P$ to 8 and define 8 window widths $[8,16,32,64,96,128,164,196]$.
We set the dimension of almost parameter matrices and bias to 256, including the ${\bf W}^{exp}_{(\cdot)}$, ${\bf b}^{exp}_{(\cdot)}$ in the explicit graph convolution, ${\bf W}^{f}$ and ${\bf b}^{f}$ in the temporal localizer and so on.  And the BiGRU networks have 128-d hidden states for each direction. During training, we apply an Adam optimizer~\cite{duchi2011adaptive} to minimize the multi-task loss ${\mathcal L}_{STGRN}$, where the initial learning rate is set to 0.001 and the batch size is 16.

\textbf{Evaluation Criteria.}
We employ the \textbf{m\_tIoU}, \textbf{m\_vIoU} and \textbf{vIoU@R} as evaluation criteria~\cite{gao2017turn,chen2019weakly}. 
The \textbf{m\_tIoU} is the average temporal IoU between the selected clips and ground truth clips. And we define ${\mathcal S}_U$ as the set of frames contained in the selected or ground truth clips, and ${\mathcal S}_I$ as the set of frames in both selected and ground truth clips. We calculate vIoU by $ {\rm vIoU} = \frac{1}{|{\mathcal S}_U|} \sum_{t \in {\mathcal S}_I} {\rm IoU}(r^t, {\hat r}^t)$, where $r^t$ and ${\hat r}^t$ are selected and ground truth regions of frame $t$. The \textbf{m\_vIoU} is the average vIoU of samples and \textbf{vIoU@R} is the proportion of samples which vIoU $>$ R.

\textbf{Baseline.}
Since no existing strategy can be directly applies to STVG, we extend the existing visual grounding method \textbf{GroundeR}~\cite{rohrbach2016grounding} and video grounding approaches \textbf{STPR}~\cite{yamaguchi2017spatio} and \textbf{WSSTG}~\cite{chen2019weakly} as the baselines. Considering these methods all lack temporal grounding, we first apply the temporal sentence localization methods \textbf{TALL}~\cite{gao2017tall} and \textbf{L-Net}~\cite{chen2019localizing} to obtain a clip and then retrieve the tubes from the trimmed clip by GroundeR, STPR and WSSTG.
The GroundeR is a frame-level approach, which originally grounds natural language in a still image. We apply it for each frame of the clip and generate a tube. The STPR and WSSTG are both tube-level methods and adopt the tube pre-generation framework. Specifically, the original STPR~\cite{yamaguchi2017spatio} only grounds persons from multiple videos, we extend it to multi-type object grounding in a single clip. The original WSSTG~\cite{chen2019weakly} employs a weakly-supervised setting, we extend it by applying a supervised triplet loss~\cite{yang2019cross} to select candidate tubes. So we obtain 6 combined baselines \textbf{GroundeR+TALL}, \textbf{STPR+TALL} and so on. 
We also provide the temporal ground truth to form 3 baselines. We show more baseline details in the supplementary material.

\subsection{Experiment Results}
Table~\ref{table:mainexp} shows the overall experiment results of all methods, where \textbf{STGRN(Greedy)} uses the greedy region selection for the tube generation rather than the dynamic method. The \textbf{Random}  selects the temporal clip and spatial regions randomly. \textbf{Tem. Gt} means that the temporal ground truth is provided. We can find several interesting points:
\begin{itemize}

\item The GroundeR+$\{\cdot\}$ methods independently ground sentences in every frame and achieve worse performance than STPR+$\{\cdot\}$ and  WSSTG+$\{\cdot\}$ methods, validating the temporal object dynamics across frames are vital for spatio-temporal video grounding.
\item The model performance on interrogative sentences is obviously lower than declarative sentences, which shows the interrogative sentences with unknown objects are more difficult to ground.
\item For temporal grounding, our STGRN achieves a better performance than the frame-level localization method TALL and L-Net, demonstrating the spatio-temporal region modeling is effective to determine the temporal boundaries of object tubes.
\item For spatio-temporal grounding, our STGRN outperforms all baselines on both declarative and interrogative sentences with or without temporal ground truth, which suggests our cross-modal spatio-temporal graph reasoning can effectively capture the object relationships with temporal dynamics and our spatio-temporal localizer can retrieve the object tubes precisely.
\item Our STGRN with the dynamic selection method outperforms the STGRN(Greedy) with the greedy method, showing the dynamic smoothness is beneficial to generate high-quality tubes.
\end{itemize}

\begin{table}[t]
\centering
\caption{Ablation Results on the VidSTG Dataset.}\label{table:ablation}
\scalebox{0.95}{
\begin{tabular}{ccc|ccc}
\hline
${\mathcal G}_{imp}$  &${\mathcal G}_{exp}$ & ${\mathcal G}_{tem}$&  m\_tIoU& m\_vIoU &vIoU@0.3\\
\hline
\checkmark&&&44.81\%&17.13\%&21.08\%\\
&\checkmark&&45.56\%&17.58\%&21.49\%\\
&&\checkmark&45.12\%&17.53\%&21.91\%\\
\checkmark&\checkmark&&45.99\%&17.72\%&22.07\%\\
\checkmark&&\checkmark&46.70\%&18.07\%&22.23\%\\
&\checkmark&\checkmark&47.12\%&18.28\%&22.52\%\\
\checkmark&\checkmark&\checkmark&{\bf 47.64\%}&{\bf 18.96\%}&{\bf 23.19\%}\\
\hline
\end{tabular}
}
\end{table}

\subsection{Ablation Study}
In this section, we conduct the ablation studies on the spatio-temporal region graph that is the key component of our STGRN. Concretely, the spatio-temporal graph includes the implicit spatial subgraph ${\mathcal G}_{imp}$ , explicit spatial subgraph ${\mathcal G}_{exp}$ and temporal dynamic subgraph ${\mathcal G}_{tem}$. We selectively discard them to generate ablation models and report all ablation results in Table~\ref{table:ablation}, where we do not distinguish the declarative and interrogative sentences. From these results, we can find that the full model outperforms all ablation models, validating each subgraph is helpful for spatio-temporal video grounding.
If only a subgraph is applied, the model with ${\mathcal G}_{exp}$ achieves the best performance, demonstrating the explicit modeling is the most important to capture object relationships. And if two subgraphs are used, the model with ${\mathcal G}_{exp}$ and ${\mathcal G}_{tem}$ outperforms other models, which suggests the spatio-temporal modeling play a crucial role in relation understanding and high-quality video grounding.

\begin{figure}[t]
\centering
\includegraphics[width=0.23\textwidth]{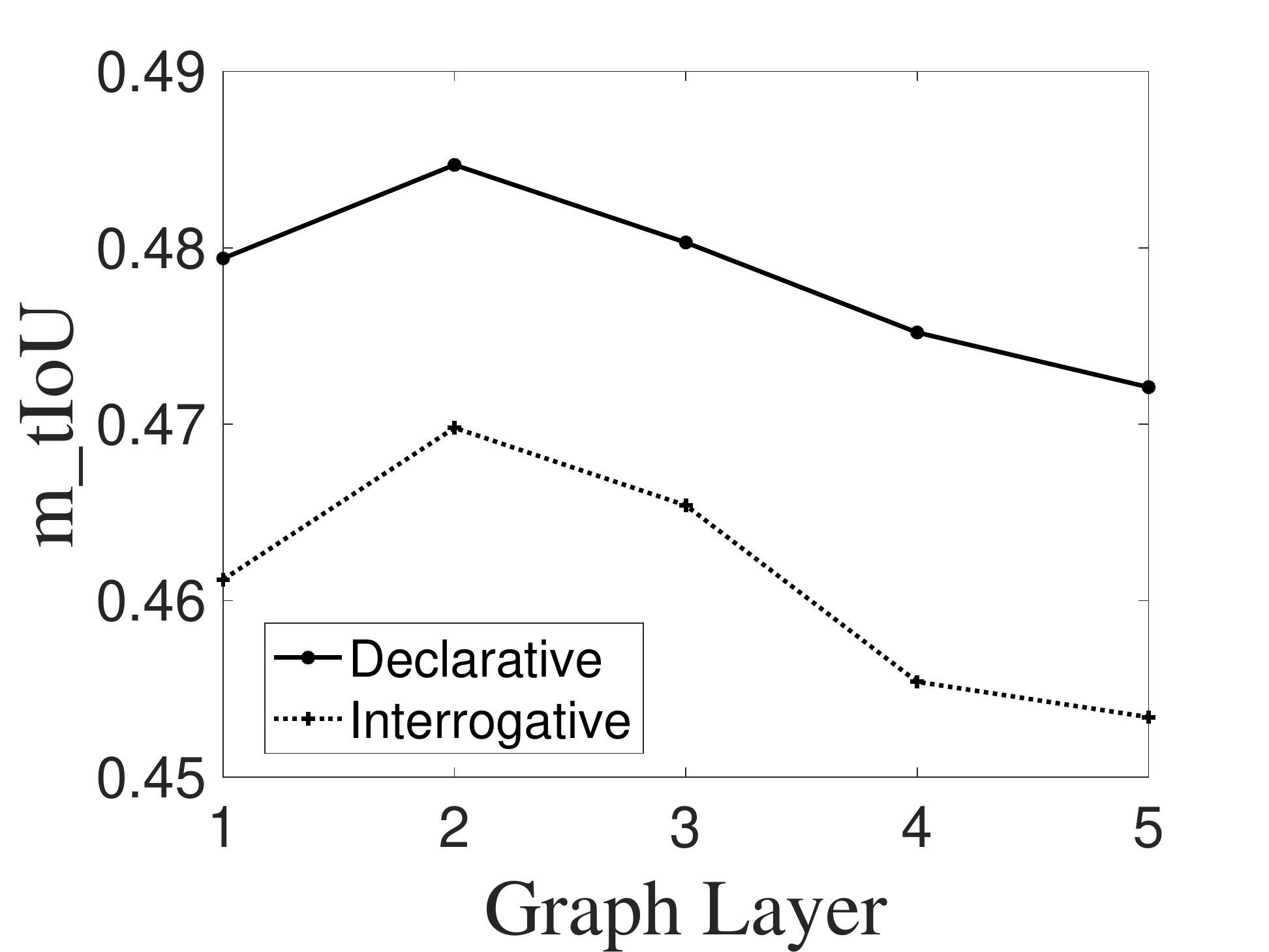}
\includegraphics[width=0.23\textwidth]{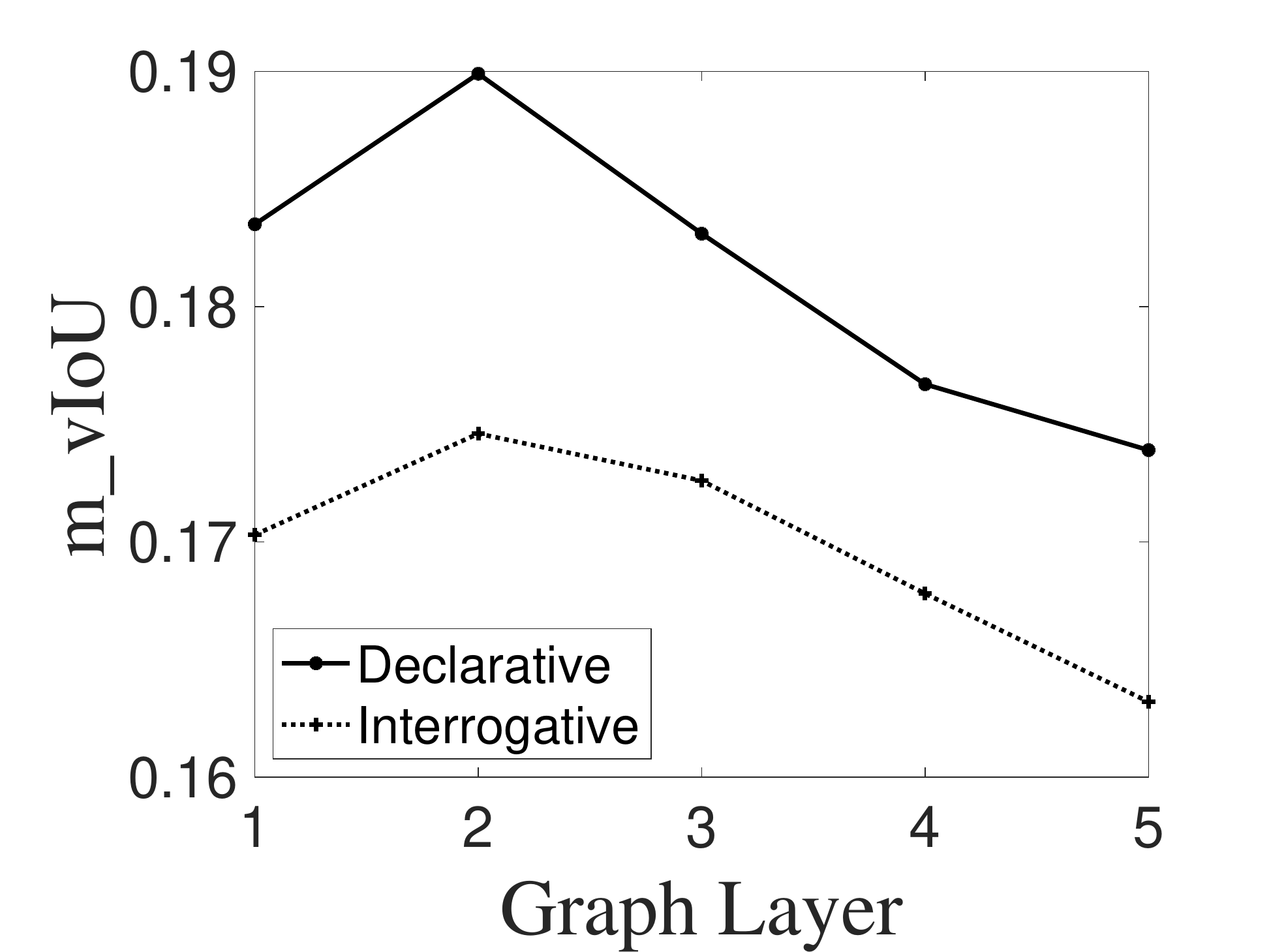}
\caption{Effect of the Number of Spatio-Temporal Graph Convolution Layers.}\label{fig:hyper}
\end{figure}

Moreover, the layer number $T$ is the essential hyperparameter of the spatio-temporal graph. We investigate the effect of $T$ by varying it from 1 to 5. Figure~\ref{fig:hyper} shows the experimental results on the criteria m\_tIoU and m\_vIoU for both declarative and interrogative sentences. From the results, we can find our STGRN has the best performance when $T$ is set to 2. The one-layer graph cannot sufficiently capture the object relationships and temporal dynamics. And too many layers may result in region over-smoothing, that is, each region feature tends to be identical.
The performance changes on different criteria and sentence types are basically consistent, demonstrating the stable influence of $T$.

\begin{figure}[t]
\centering
\includegraphics[width=0.48\textwidth]{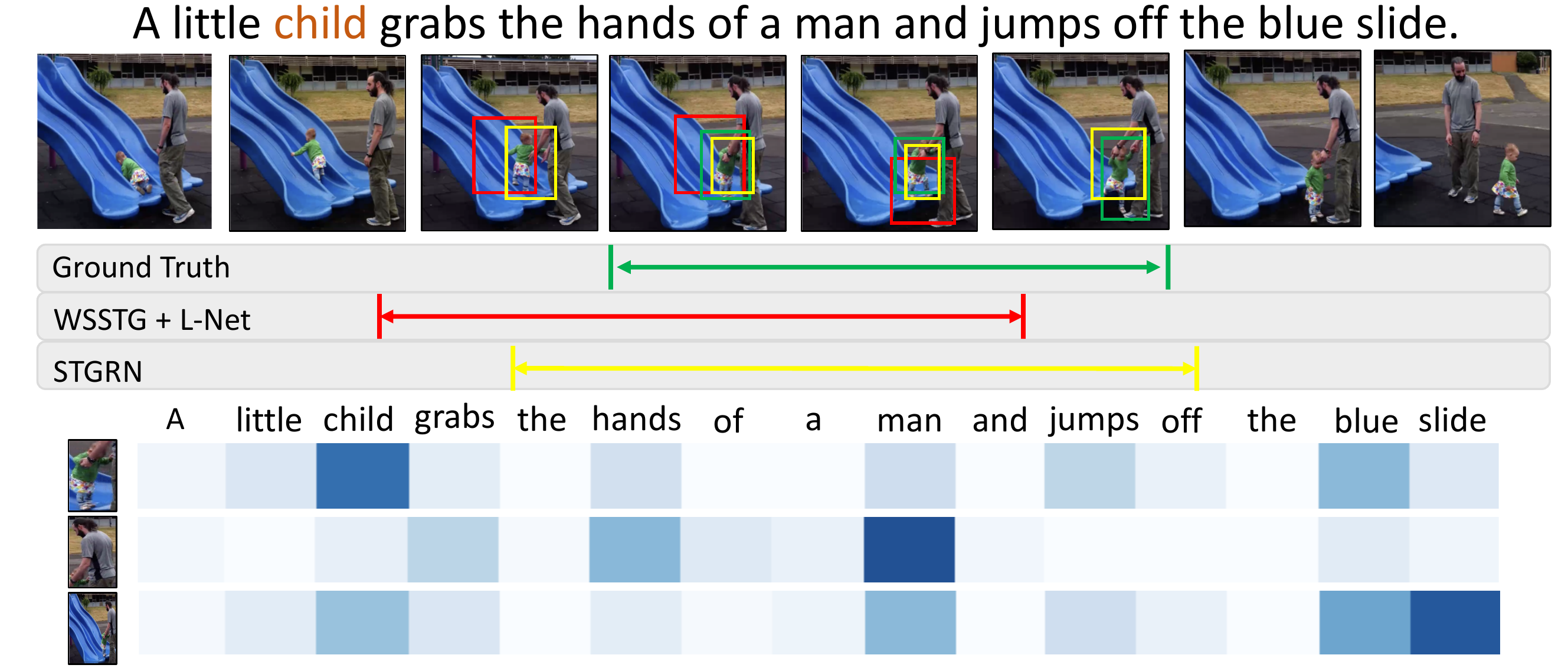}
\caption{A Example of the Spatio-Temporal Grounding Result.}\label{fig:exampleres}
\end{figure}

\subsection{Qualitative Analysis}
We display a typical example in Figure~\ref{fig:exampleres}. The sentence describes two parallel actions "grab the hands" and "jump off the slide" of the boy in a short-term segment, requiring the accurate spatio-temporal grounding. By intuitive comparison, our STRGN gives a more precise temporal segment and generates a more reasonable spatio-temporal tube than the baseline WSSTG+L-net.
Moreover, the attention method in the cross-modal fusion module builds a bridge between visual and textual contents, and we visualize the weights of several key regions over the sentences here. We can see that semantic related region-word pairs have a larger weight, e.g., the region of the boy and the word "child".

\section{Conclusion}
In this paper, we propose a novel spatio-temporal video grounding task STVG and contribute a large-scale dataset VidSTG. 
We then design a STGRN to capture region relationships with temporal object dynamics and directly localize the spatio-temporal tubes from the region level. 

\noindent \textbf{Acknowledgments}
This work was supported by Zhejiang Natural Science Foundation LR19F020006 and the National Natural Science Foundation of China under Grant No.61836002, No.U1611461 and No.61751209, Sponsored by China Knowledge Centre for Engineering Sciences and Technology and Alibaba-Zhejiang University Joint Research Institute of Frontier Technologies.


{\small
\bibliographystyle{ieee_fullname}
\bibliography{ref}
}

\newpage
\twocolumn[{%
\maketitle
\centering
\section*{Where Does It Exist: Spatio-Temporal Video Grounding for Multi-Form Sentences \\--Supplementary Material--}
\vspace{1em}
}]

\section{Dataset Details}

VidOR contains 7,000, 835 and 2,165 videos for training, validation and testing, respectively. Since box annotations of testing videos are unavailable yet, we omit testing videos, split 10\% training videos as our validation data and regard original validation videos as the testing data. Considering a video may contain multiple same triplets that have different temporal and bounding box annotations, we cut these videos into several short videos, where each short video contains a triplet that covers a segment of the short video.
We then delete unsuitable video-triplet pairs based on three rules: (1) the video length is less than 3 seconds; (2) the temporal duration of the triplet is less than 0.5 seconds; (3) the triplet duration is less than 2\% of the video. Next, because too many triplets are related to spatial relations like "in\_front\_of" and "next\_to", we delete 90\% spatial triplets to keep the types of relations balanced. 

\begin{figure*}[t]
\centering
\includegraphics[width=0.98\textwidth]{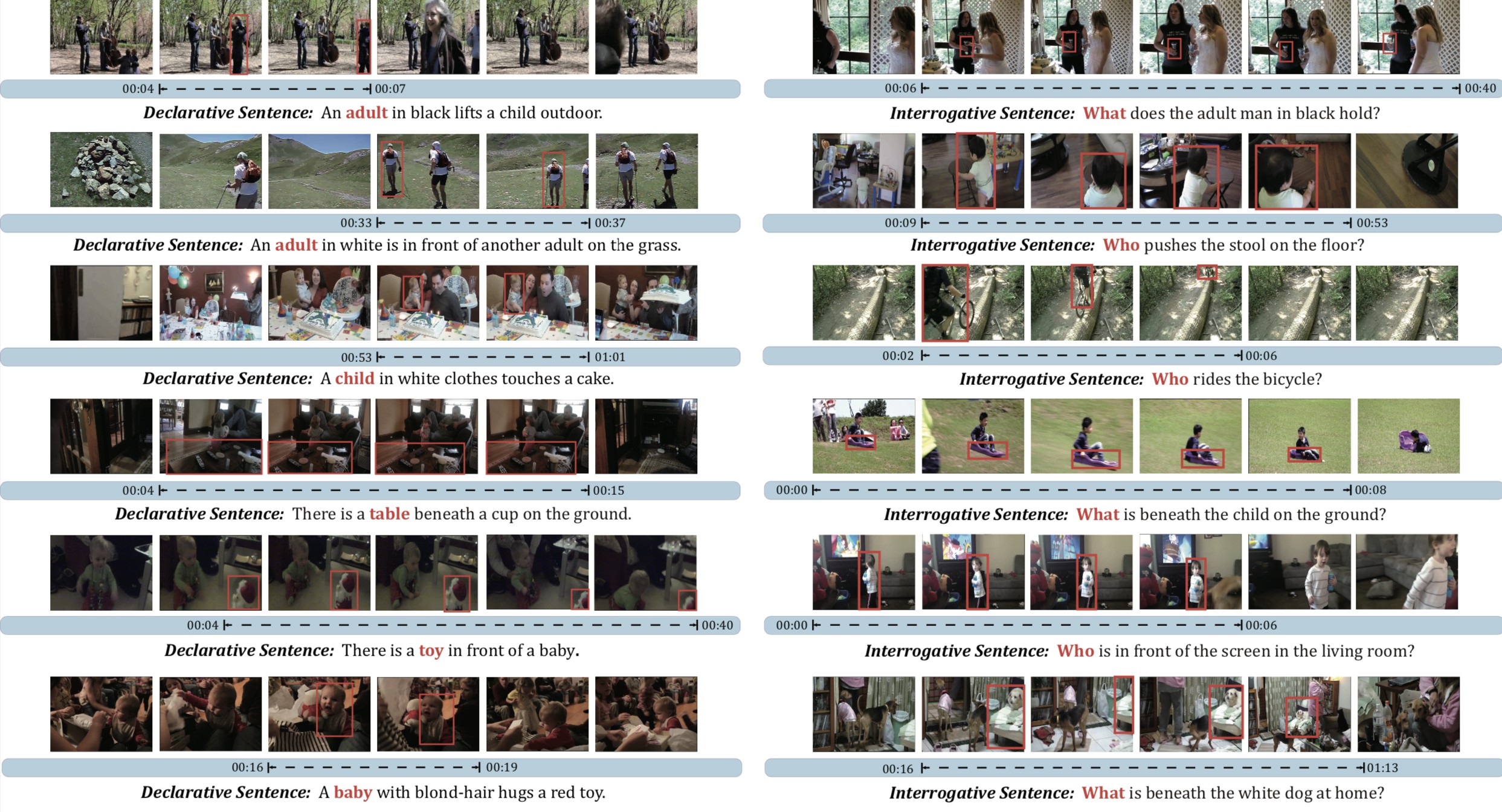}
\caption{Annotation Samples with Declarative or Interrogative Sentence Descriptions. }\label{fig:samples}
\end{figure*}

For each video-triplet pair, we choose the $subject$ or $object$ as the queried object, and then describe its appearance, relationships with other objects and visual environments. We discard video-triplet pairs that are too hard to give a precise description. And a video-triplet pair may correspond to multiple sentences.
After annotation, there are 6,924 videos (5,563, 618 and 743 for training, validation and testing sets) and 99,943 sentences for 44,808 video-triplet pairs. 
we show some typical samples in Figure~\ref{fig:samples} with declarative and interrogative sentences. We can find that the objects may exist in a very small segment of the video and the sentences may only describe a short-term state of the queried object.

\begin{figure*}[t]
\centering
\includegraphics[width=0.98\textwidth]{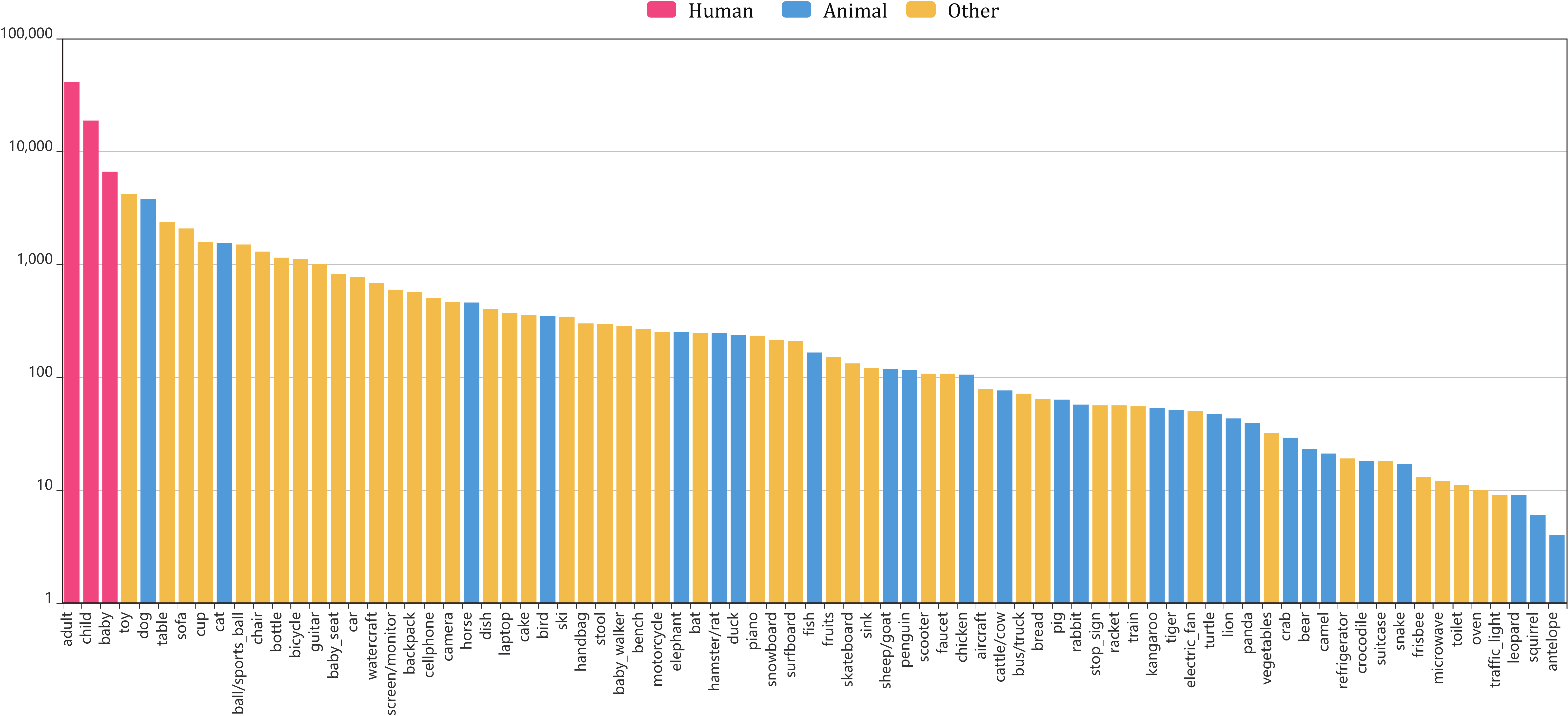}
\caption{The Distribution of Different Types of Queried Objects in the Entire VidSTG Dataset. }\label{fig:distribution}
\end{figure*}

Next, we show the distribution of different types of queried objects as Figure~\ref{fig:distribution}. The original VidOR contains 80 types of objects, including 3 types of persons, 28 types of animals and 49 types of other objects. After data cleaning and annotating, sentences in VidSTG describes 79 types of objects by the declarative or interrogative ways, including 3 types of persons, 27 types of animals and 49 types of other objects. A rare type (i.e., stingray) is not contained in VidSTG. From Figure~\ref{fig:distribution}, we can find the sentences of person types take up the largest proportion and sentence numbers of other categories are relatively uniform. 

Moreover, we compare VidSTG with existing video grounding datasets in Table~\ref{table:dataset}. 
Previous temporal sentence grounding datasets like DiDeMo~\cite{hendricks2017localizing}, Charades-STA~\cite{gao2017tall}, TACoS~\cite{regneri2013grounding} and ActivityCation~\cite{krishna2017dense} only provide the temporal annotations for each sentence and lack the spatio-temporal bounding boxes. 
As for existing video grounding datasets, Persen-sentence~\cite{yamaguchi2017spatio} is originally used for spatio-temporal person retrieval among trimmed videos and only contains one type of objects (i.e. people), which is too simple for the STVG task. And VID-sentence dataset~\cite{chen2019weakly} contains 30 categories but also offer the annotations on trimmed videos. 
Different from them, our VidSTG simultaneously offers temporal clip and spatio-temporal tube annotations, contains more sentence descriptions, has a richer variety of objects, and further supports multi-form sentences.

\begin{table*}[t]
\centering
\caption{Dataset Comparison.}\label{table:dataset}
\begin{tabular}{c|cccccccc}
\hline
Dataset& \#Video & \#Sentence &\#Type& Domain &Temporal Ann.&Box Ann.& Multi-Form Sent.  \\
\hline
TACoS&127&18,818&-&Cooking&\checkmark&&\\
DiDeMo&10,464&40,543&-&Open&\checkmark&&\\
Charades-STA&6,670&16,128&-&Activity&\checkmark&&&\\
ActivityCaption&20,000&37,421&-&Open&\checkmark&&\\
Person-sentence&5,293&30,365&1&Person&&\checkmark&\\
VID-sentence&4,318&7,654&30&Open&&\checkmark&\\
\hline
VidSTG&6,924&99,943&79&Open&\checkmark&\checkmark&\checkmark\\
\hline
\end{tabular}
\end{table*}

\section{Baseline Details}
Since no existing strategy can be directly applies to STVG, we combine the existing visual grounding method \textbf{GroundeR}~\cite{rohrbach2016grounding} and spatio-temporal video grounding approaches \textbf{STPR}~\cite{yamaguchi2017spatio} and \textbf{WSSTG}~\cite{chen2019weakly} with the temporal sentence localization methods \textbf{TALL}~\cite{gao2017tall} and \textbf{L-Net}~\cite{chen2019localizing} as the baselines. The TALL and L-Net first provide the temporal clip of the target tube and the extended GroundeR, STPR and WSSTG then retrieve the spatio-temporal tubes of objects.

We first introduce the TALL and L-Net approaches. The TALL applies a sliding window framework that first samples abundant candidate clips and then ranks them by estimating the clip-sentence scores. During estimating, TALL incorporates the context features for the current clip to further improve the localization accuracy. 
And L-Net develops the evolving frame-by-word interactions for video and query contents, and dynamically aggregates the matching evidence to localize the temporal boundaries of clips according to the textual query.

Next, we illustrate the extended grounding methods GroundeR, STPR and WSSTG based on the retrieved clip. The GroundeR is a frame-level approach, which originally grounds natural language in a still image. We apply it for each frame of the clip to obtain the object region and generate a tube by directly connecting these regions. The drawback of this method is the lack of temporal context modeling of regions.
Different from it, original STPR and WSSTG are both tube-level methods and adopt the tube pre-generation framework.
This framework first extracts a set of spatio-temporal tubes from trimmed clips and then identifies the target tube. 
The original STPR~\cite{yamaguchi2017spatio} only grounds persons from multiple videos, we extend it to multi-type object grounding in a single clip. Specifically, we use the pre-trained Faster R-CNN to detect multi-type object regions to generate the candidate tubes rather than only generate person candidate tubes. And during training, we retrieve the correct tube from a video rather multiple videos, where we do not change the loss function of STPR.
The original WSSTG~\cite{chen2019weakly} employs a weakly-supervised setting, we extend it to the fully-supervised form.
Concretely, we discard the original ranking and diversity losses and employs a classics triplet loss~\cite{yang2019cross} on the matching scores of the candidate tubes and sentence. The STPR and WSSTG both have the drawbacks of the tube pre-generation framework: (1) they are hard to pre-generate high-quality tubes without textual clues; (2) they only consider single tube modeling and ignore the rela1ionships between objects.
Finally, we obtain 6 combined baselines \textbf{GroundeR+TALL}, \textbf{STPR+TALL}, \textbf{WSSTG+TALL}, \textbf{GroundeR+L-Net}, \textbf{STPR+L-Net} and \textbf{WSSTG+L-Net}.
We also provide the temporal ground truth clip to form 3 baselines \textbf{GroundeR+Tem.Gt}, \textbf{STPR+Tem.Gt} and \textbf{WSSTG+Tem.Gt}.

During training, we first train the TALL and L-Net based on the sentence-clip matching data and train GroundeR, STPR and WSSTG within the ground truth clip. But while inference, we first use TALL and L-Net to determine the clip boundaries and then employ GroundeR, STPR and WSSTG to localize the final tubes.
To guarantee the fair comparison,  GroundeR, STPR and WSSTG are built on the same TALL or L-Net models, and we apply the Adam optimizer to train all baselines.

\begin{table}[t]
\centering
\caption{Ablation Results of Directed GCN.}\label{table:ablation1}
\scalebox{0.75}{
\begin{tabular}{c|cccc}
\hline
Method&  m\_tIoU& m\_vIoU &vIoU@0.3 &vIoU@0.5\\
\hline
w/o. Explicit Subgraph &46.70\%&18.07\%&22.23\%&13.12\%\\
Undirected GCN &46.81\%&18.13\%&22.28\%&13.06\%\\
Undirected GAT &47.08\%&18.21\%&22.35\%&13.20\%\\
Directed GCN~(our)&{\bf 47.64\%}&{\bf 18.96\%}&{\bf 23.19\%}&{\bf 13.62\%}\\
\hline
\end{tabular}
}
\end{table}

\begin{table}[t]
\centering
\caption{Ablation Results of Query Modeling.}\label{table:ablation2}
\scalebox{0.7}{
\begin{tabular}{c|c|ccc}
\hline
Method& Query Modeling& m\_tIoU& m\_vIoU &vIoU@0.3\\
\hline
\multirow{2}{*}{WSSTG+L-Net} &GRU &40.27\%&13.85\%&17.66\%\\
&GRU+Object Rec.+Attention &41.22\%&14.32\%&20.08\%\\
\hline
\multirow{2}{*}{STGRN} &GRU &46.93\%&18.42\%&22.41\%\\
&GRU+Object Rec.+Attention&{\bf 47.64\%}&{\bf 18.96\%}&{\bf 23.19\%}\\
\hline
\end{tabular}
}
\end{table}

\section{More Ablation Study}
\subsection{Directed GCN}
To confirm the effect of the directed explicit GCN, we replace it with the original undirected GCN~\cite{kipf2016semi} and GAT~\cite{velivckovic2017graph}. In Table~\ref{table:ablation1}, our directed GCN has a better performance and the results of undirected GCN and GAT are close to the model without explicit subgraph modeling. The reason is that the undirected GCN and GAT have a similar ability with the implicit GCN and may lead to redundancy modeling.

\subsection{Query Modeling}
Our input setting is consistent with previous grounding works but we adopt a different strategy for query modeling in STGRN. Previous works model the sentence by RNN as a whole query vector. Different from them, we use the NLTK library to recognize the first noun or interrogative word "who/what" in the sentence, corresponding to the query object. We then select its feature ${\bf s}^e$ from RNN outputs and adopt context attention to learn the object-aware query vector ${\bf s}^q$. We conduct an ablation study for query modeling in Table~\ref{table:ablation2}, where we also apply the \textbf{GRU+Object Rec.+Attention} to WSSTG+L-Net.
Concretely, the object-aware vector ${\bf s}^q$ replaces the original sentence vector in final localization for L-Net and is added into visually guided sentence features for WSSTG.

\end{document}